\documentclass[acmtog, submission, ]{acmart}

\usepackage{wrapfig}

\usepackage{wrapfig}

\copyrightyear{2022}
\acmYear{2022}
\setcopyright{rightsretained}
\acmConference[NICE 2022]{Neuro-Inspired Computational Elements Conference}{March 28-April 1, 2022}{Virtual Event, USA}
\acmBooktitle{Neuro-Inspired Computational Elements Conference (NICE 2022), March 28-April 1, 2022, Virtual Event, USA}\acmDOI{10.1145/3517343.3517367}
\acmISBN{978-1-4503-9559-5/22/03}

\begin{document}

\title{Sequence Learning and Consolidation on Loihi using On-chip Plasticity}

\author{Jack Lindsey}
\email{jackwlindsey@gmail.com}
\affiliation{%
  \institution{Columbia University}
  \city{New York}
  \state{NY}
  \country{USA}}

\author{James B Aimone}
\affiliation{%
  \institution{Sandia National Laboratories}
  \city{Albuquerque}
  \state{NM}
  \country{USA}}
\email{jbaimon@sandia.gov}

\renewcommand{\shortauthors}{Lindsey and Aimone}
\begin{abstract}
In this work we develop a model of predictive learning on neuromorphic hardware.  Our model uses the on-chip plasticity capabilities of the Loihi chip to remember observed sequences of events and use this memory to generate predictions of future events in real time.  Given the locality constraints of on-chip plasticity rules, generating predictions without interfering with the ongoing learning process is nontrivial.  We address this challenge with a memory consolidation approach inspired by hippocampal replay.  Sequence memory is stored in an initial memory module using spike-timing dependent plasticity.  Later, during an offline period, memories are consolidated into a distinct prediction module.  This second module is then able to represent predicted future events without interfering with the activity, and plasticity, in the first module, enabling online comparison between predictions and ground-truth observations.  Our model serves as a proof-of-concept that online predictive learning models can be deployed on neuromorphic hardware with on-chip plasticity.
\end{abstract}

\begin{teaserfigure}
  \includegraphics[width=\textwidth,trim={0 7cm 0 4.7cm},clip]{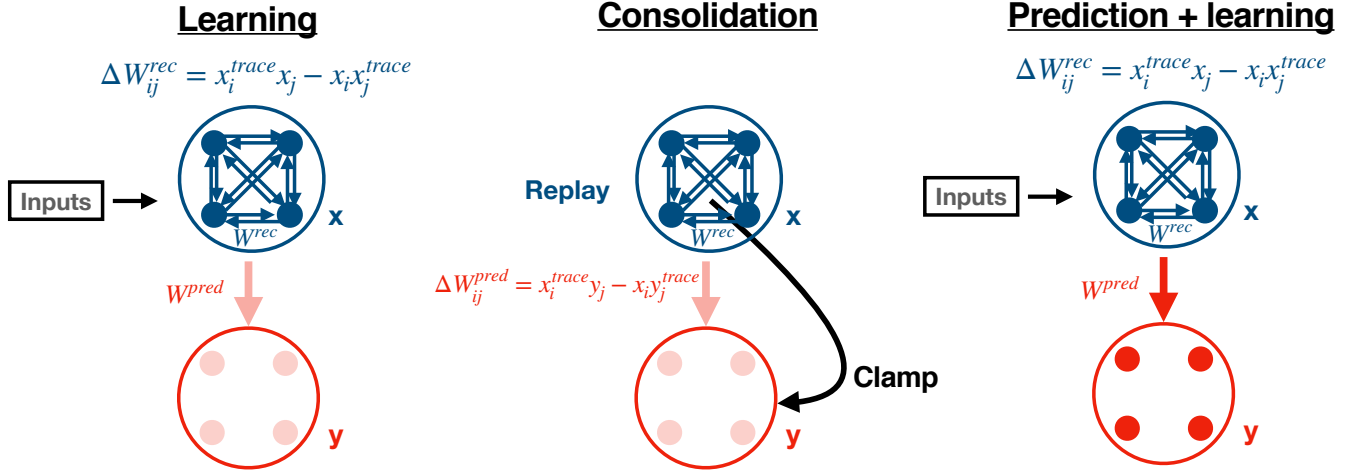}
  \caption{Model schematic, illustrating separate online learning and offline consolidation phases.}
  \label{fig:teaser}
\end{teaserfigure}

\maketitle

\section{Introduction}
Neuromorphic hardware is attracting growing interest for its potential to enhance energy efficiency compared to conventional computing.  However, the hardware features that enable these efficiency improvements -- namely, spike-based communication and local information processing -- constrain the space of algorithms that can be implemented.  Developing effective methods to train spiking neural networks for deployment on neuromorphic hardware is an active area of research \cite{zenke2021brain, cramer2020training}.  Here we focus on a relatively less-studied issue: the use of on-chip plasticity for online learning and adaptation post-deployment.  While prior work has explored approaches to on-chip associative learning \citep{imam2020rapid, hampo2020associative}, we tackle the problem of learning sequential structure in the environment and using it to make predictions about future events.  We make use of Intel's Loihi chip, which allows the use of a rich family of user-defined on-chip plasticity rules \cite{davies2018loihi}.  We suggest and implement a biologically inspired sequence learning algorithm that makes use of an offline experience replay protocol to consolidate learned information into a separate module, which can then be used to generate predictions without interfering with ongoing learning. The goal of this research is to provide a compact online sequence learning model that is compatible with learning neuromorphic systems. We believe that introducing such a model to the community will facilitate the development of novel neuromorphic algorithms designed to leverage on-chip learning as well as providing a concrete algorithm for benchmarking and evaluating proposed neuromorphic architectures.


\begin{figure}
  \includegraphics[width=0.45\textwidth]{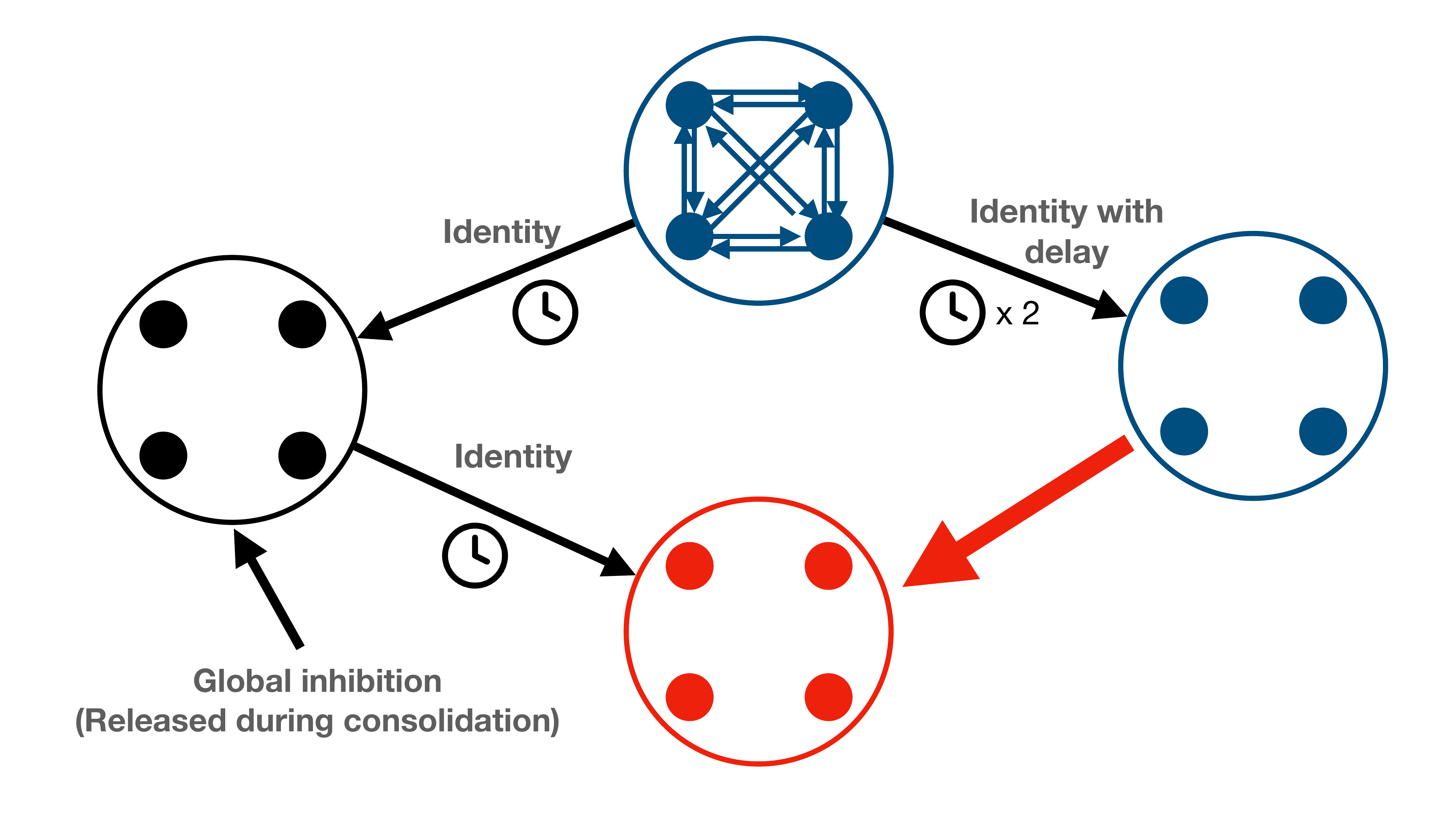}
  \caption{Implementation schematic illustrating auxiliary neural populations with delayed communication.}
  \label{fig:implementation}
\end{figure}

\section{Task}.
We consider an abstract environment in which an agent is exposed to a stream of pattern sequences $S_i = [p_i^1, ..., p_i^L]$ of length $L$.  Patterns $p_i^j$ activate a particular sparse (fraction $1/8$) subset $x_i^j$ of a population (the ``sensory module'') of spiking neurons of size $N = 128.$.  The agent is tasked with learning the structure of the sequences, such that when $x_i^j$ is presented, the agent activates a set of neurons $y_i^{j+1}$ in a distinct population of $N$ neurons -- the ``prediction module'' -- that corresponds to $p_i^{j+1}$.  We assume that the neurons of the prediction population $y$ are in one-to-one correspondence with the first population. For terminal patterns $p_i^L$ we would like the prediction module to be silent.

\section{Approach and results} 
Fig. \ref{fig:teaser} illustrates the approach described below.

\emph{Online learning phase}.  In the sensory module neurons, which are connected via recurrent synapses $W^{rec}$, we employ a plasticity rule of the form $\Delta W_{ij} = r(x_i^{trace} x_j - x_i x_j^{trace})$, where the $x_i$ are indicator variables for a spike in neuron $i$ at the present timestep,  the $x_i^{trace}$ variables  are eligibility traces computed as an exponentially decaying cumulative tally of recent spikes.  Here $r$ is a global modulatory variable that is set to 1 during online learning, but set to 0 in a later consolidation phase (see below).  This rule corresponds to spike timing-dependent plasticity (STDP) and tends to strengthen weights $W_{ij}$ when neuron $i$ tends to spike prior to neuron $j$ at the timescale of the eligibility trace decay.

\begin{figure}
  \includegraphics[width=0.45\textwidth]{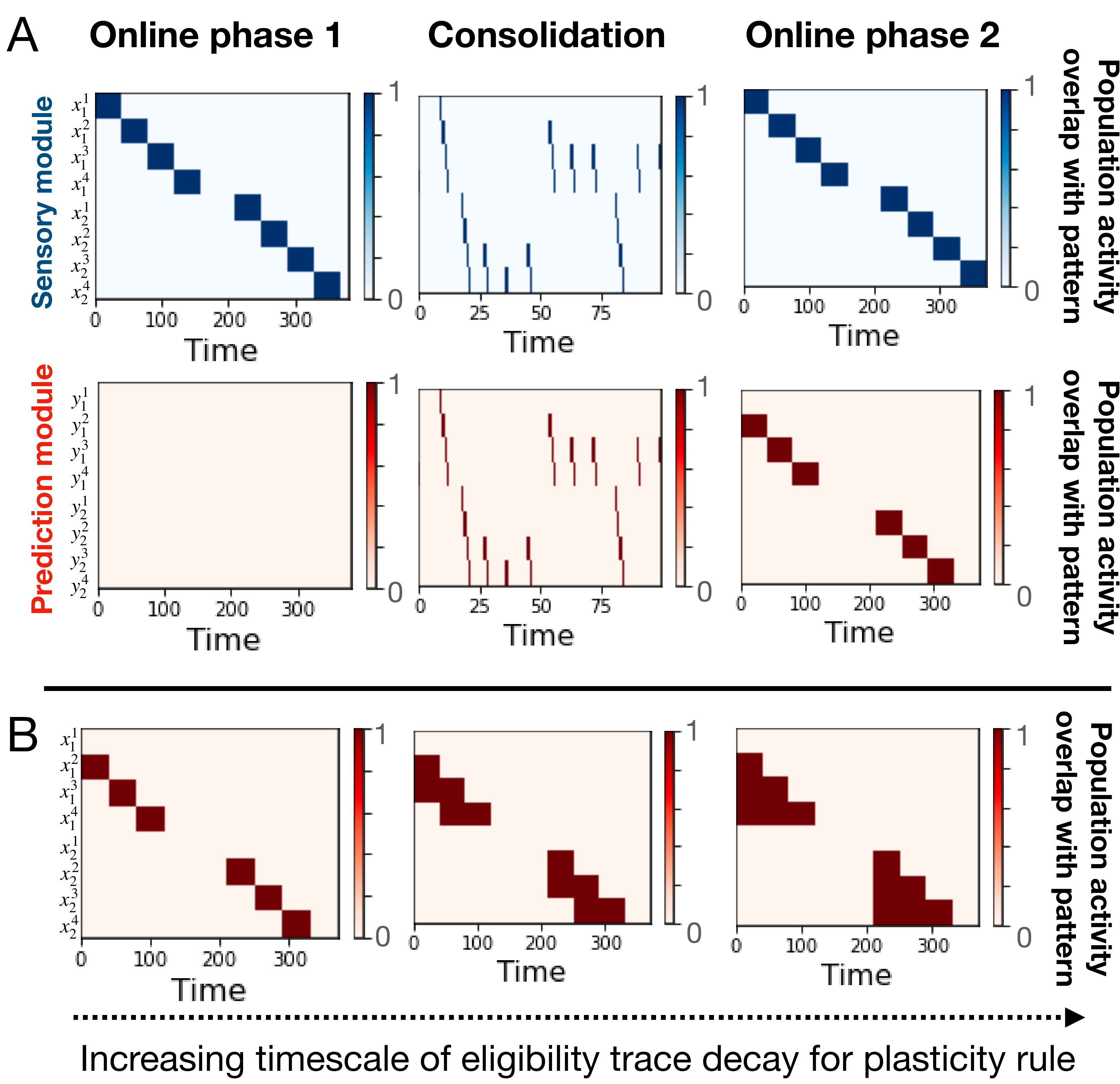}
  \caption{A. Top row: In the sensory module, overlap of the spiking neural activity at each time step with the responses evoked by each pattern (y axis), defined as the fraction of pattern-evoked neurons that are active.  During the online phases, these overlaps reflect sensory inputs faithfully.  During the consolidation period, they reflect rapid replay of sequences experienced during the online phase.  Bottom row: same as above, for the neurons in the prediction module. Prior to learning, no activity is evoked in the prediction module during the online phase.  During consolidation, activity mimicks that of the sensory module.  Following consolidation, the prediction module reperesents predicted next patterns. B. Illustration of the effect of varying the time constant of the STDP plasticity rule.  A longer time constant enables longer-term predictions.}
  \label{fig:results}
\end{figure}

\emph{Consolidation phase}.  We employ a memory consolidation strategy inspired by replay activity in the hippocampus, which plays a key role in sequence learning and memory \citep{poldrack2003sequence, foster2012sequence}.  The hippocampus is known to consolidate recently experienced sequences by replaying the activity patterns they induced \cite{carr2011hippocampal}.  In an offline period, we use a consolidation strategy reactivate randomly sampled patterns in the sensory module and allow its dynamics to run freely.  Due to the STDP-driven learning, this has the effect of replaying observed sequences of patterns in the sensory module's activity.  During this phase, we allow the sensory module neurons to drive the corresponding neurons in the prediction module via special one-to-one interactions, which are inhibited during online learning and disinhibited during consolidation (additional details below).  Aside from these connections, the  sensory and prediction module neurons are connected in all-to-all fashion by weights $W_{pred}$, which undergoes plasticity according to rule $\Delta W_{pred}^{ij} = (1-r) x_i^{trace} y_j - x_i y_j^{trace}$.  Note the factor $1-r$, which ensures that $W^{pred}$ is only modified during the consolidation phase.  This process has the effect of encoding the same predictive structure in the weights $W^{pred}$ that was initially learned in $W^{rec}$.  In this fashion, the learned information is consolidated to a separate set of synapses and neurons, freeing the sensory module to learn new associations during subsequent online learning.

\emph{Auxiliary populations and delayed connectivity}.  Gating connections directly with a modulatory term is not possible on Loihi, and hence an intermediate population must be used, which can be gated on/off during the consolidation/online phases with inhibitory inputs.  Additionally, interactions between neurons on Loihi come with a minimum one timestep latency, which poses problems for the plasticity rule in the consolidation phase. To circumvent this issue, we introduce another auxiliary popoulation that receives delayed identity connections from the sensory module, such that during the consolidation phase its activity is synchronized with that of the prediction module. See Fig. \ref{fig:implementation} for an illustration.

\emph{Results}. See Fig. \ref{fig:results}A for an illustrative example of the behavior of the system.  We may also desire the system to make predictions at longer timescales, rather than predict only the next pattern.  By modulating the time scale of the STDP learning rule, we can control whether spiking activity in the prediction module reflects a prediction of the next pattern or the next several patterns (Fig. \ref{fig:results}B).

\bibliographystyle{ACM-Reference-Format}
\bibliography{sample-base}

\end{document}